\documentclass[sigconf]{acmart}

\usepackage{booktabs} 
\usepackage{mathrsfs}
\usepackage{multirow}
\usepackage{graphicx}
\usepackage{epsfig}
\usepackage{graphics}
\usepackage{subfigure}
\usepackage{psfrag}
\usepackage{stfloats}
\usepackage{algorithm}
\usepackage{algorithmic}
\usepackage{stfloats}
\usepackage{amsmath}
\usepackage{amssymb}
\usepackage{amsfonts}
\usepackage{color}
\usepackage{booktabs}
\usepackage{slashbox}

\usepackage{pgfplots}
\usepackage{pgfplotstable}

\begin{document}
\copyrightyear{2017}
\acmYear{2017}
\setcopyright{acmcopyright}
\acmConference{MM '17}{October 23--27, 2017}{Mountain View, CA,
USA}\acmPrice{15.00}\acmDOI{10.1145/3123266.3127905}
\acmISBN{978-1-4503-4906-2/17/10}


\fancyhead{}
\settopmatter{printacmref=false, printfolios=false}

\title{To Create What You Tell: Generating Videos from Captions}
\titlenote{{\small This work was performed at Microsoft Research Asia. The first two authors made equal contributions to this work.}}

\author{Yingwei Pan, Zhaofan Qiu, Ting Yao, Houqiang Li and Tao Mei}
\affiliation{\institution{University of Science and Technology of China, Hefei, China \and Microsoft Research, Beijing, China}}
\email{{{panyw.ustc, zhaofanqiu}@gmail.com;{tiyao, tmei}@microsoft.com;lihq@ustc.edu.cn}}
%
%
%
%
%
%
%

\begin{abstract}
We are creating multimedia contents everyday and everywhere. While automatic content generation has played a fundamental challenge to multimedia community for decades, recent advances of deep learning have made this problem feasible. For example, the Generative Adversarial Networks (GANs) is a rewarding approach to synthesize images. Nevertheless, it is not trivial when capitalizing on GANs to generate videos. The difficulty originates from the intrinsic structure where a video is a sequence of visually coherent and semantically dependent frames. This motivates us to explore semantic and temporal coherence in designing GANs to generate videos. In this paper, we present a novel Temporal GANs conditioning on Captions, namely TGANs-C, in which the input to the generator network is a concatenation of a latent noise vector and caption embedding, and then is transformed into a frame sequence with 3D spatio-temporal convolutions. Unlike the naive discriminator which only judges pairs as fake or real, our discriminator additionally notes whether the video matches the correct caption. In particular, the discriminator network consists of three discriminators: video discriminator classifying realistic videos from generated ones and optimizes video-caption matching, frame discriminator discriminating between real and fake frames and aligning frames with the conditioning caption, and motion discriminator emphasizing the philosophy that the adjacent frames in the generated videos should be smoothly connected as in real ones. We qualitatively demonstrate the capability of our TGANs-C to generate plausible videos conditioning on the given captions on two synthetic datasets (SBMG and TBMG) and one real-world dataset (MSVD). Moreover, quantitative experiments on MSVD are performed to validate our proposal via Generative Adversarial Metric and human study.
\end{abstract}

%
%




\maketitle

\begin{figure}[!tb]
   \centering {\includegraphics[width=0.5\textwidth]{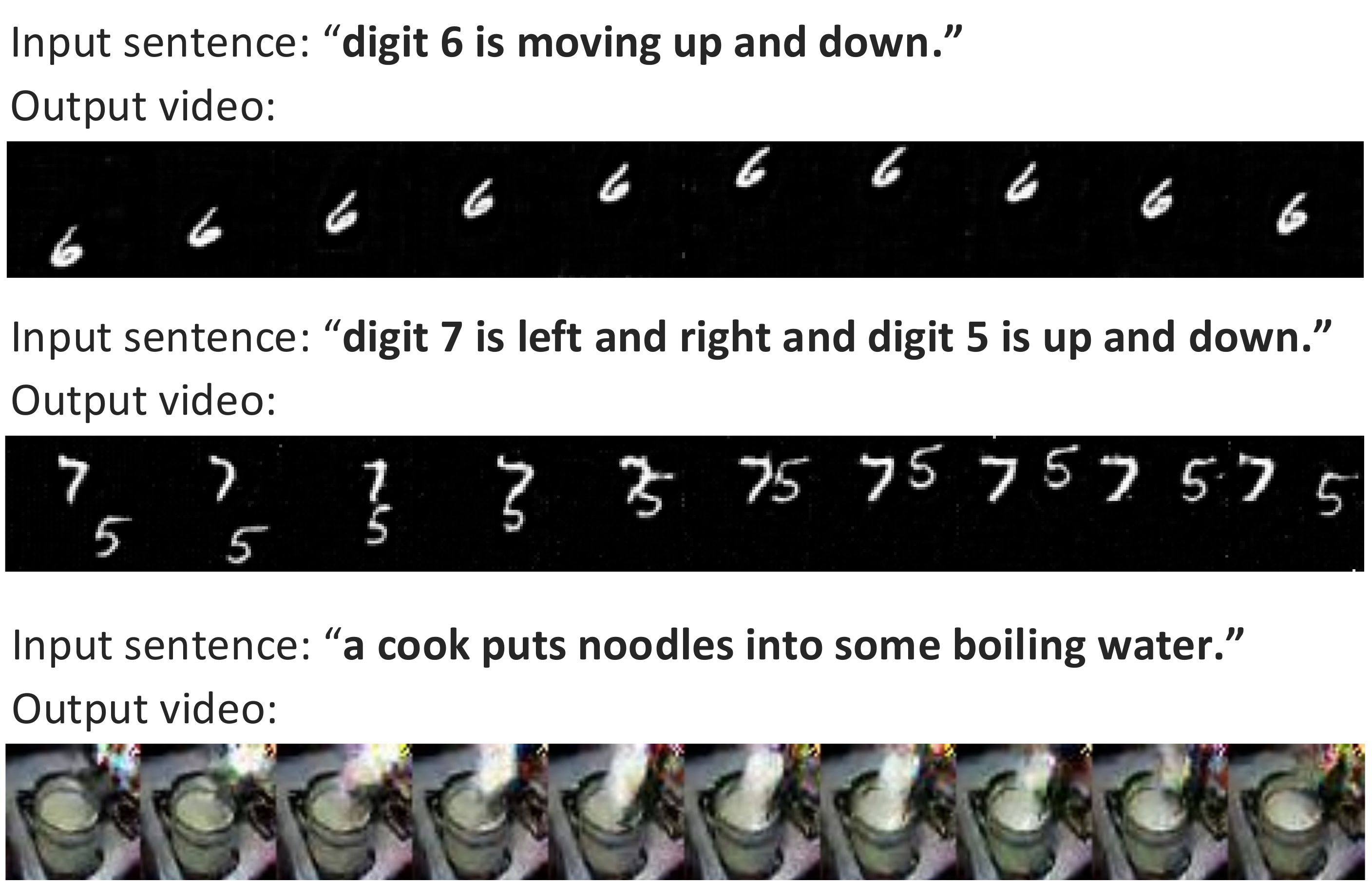}}
   \caption{\small Examples of video generation from captions on Single-Digit Bouncing MNIST GIFs, Two-Digit Bouncing MNIST GIFs and Microsoft Research Video Description Corpus, respectively.}
   \label{fig:fig1}
\end{figure}

\section{Introduction}
Characterizing and modeling natural images and videos remains an open problem in computer vision and multimedia community. One fundamental issue that underlies this challenge is the difficulty to quantify the complex variations and statistical structures in images and videos. This motivates the recent studies to explore Generative Adversarial Nets (GANs) \cite{Goodfellow:NIPS14} in generating plausible images \cite{Denton:NIPS15,Radford:ICLR16}. Nevertheless, a video is a sequence of frames which additionally contains temporal dependency, making it extremely hard to extend GANs to video domain. Moreover, as videos are often accompanied by text descriptors, e.g., tags or captions, learning video generative models conditioning on text then reduces sampling uncertainties and has a great potential real-world applications. Particularly, we are interested in producing videos from captions in this work, which is a brave new and timely problem. It aims to generate a video which is semantically aligned with the given descriptive sentence as illustrated in Figure 1.

In general, there are two critical issues in video generation employing caption conditioning: temporal coherence across video frames and semantic match between caption and the generated video. The former yields insights into the learning of generative model that the adjacent video frames are often visually and semantically coherent, and thus should be smoothly connected over time. This can be regarded as an intrinsic and generic property to produce a video. The later pursues a model with the capability to create realistic videos which are relevant to the given caption descriptions. As such, the conditioned treatment is taken into account, on one hand to create videos resembling the training data, and on the other, to regularize the generative capacity by holistically harnessing the relationship between caption semantics and video content.

\begin{figure*}[!tb]
    \centering {\includegraphics[width=1\textwidth]{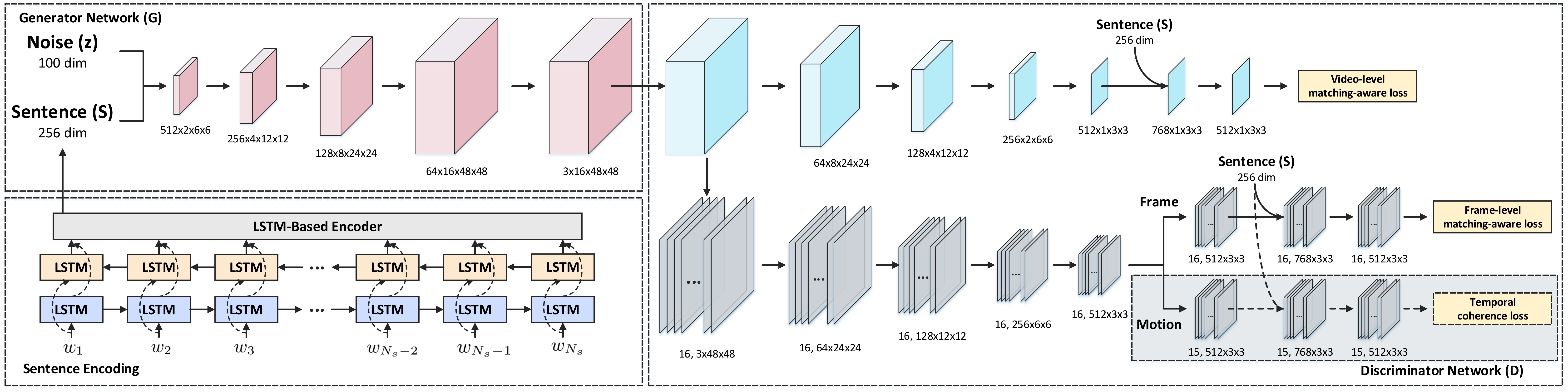}}
    \caption{\small Temporal GANs conditioning on Captions (TGANs-C) framework mainly consists of a generator network $G$ and a discriminator network $D$ (better viewed in color). Given a sentence ${\mathcal {S}}$, a bi-LSTM is first utilized to contextually embed the input word sequence, followed by a LSTM-based encoder to obtain the sentence representation ${\bf{S}}$. The generator network $G$ tries to synthesize realistic videos with the concatenated input of the sentence representation ${\bf{S}}$ and random noise variable ${\bf{z}}$. The discriminator network $D$ includes three discriminators: video discriminator to distinguish real video from synthetic one and align video with the correct caption, frame discriminator to determine whether each frame is real/fake and semantically matched/mismatched with the given caption, and motion discriminator to exploit temporal coherence between consecutive frames. Accordingly, the whole architecture is trained with the video-level matching-aware loss, frame-level matching-aware loss and temporal coherence loss in a two-player minimax game~mechanism.}
    \label{fig:fig2}
\end{figure*}

By jointly consolidating the idea of temporal coherence and semantic match in translating text in the form of sentence into videos, this paper extends the recipe of GANs and presents a novel Temporal GANs conditioning on Caption (TGANs-C) framework for video generation, as shown in Figure \ref{fig:fig2}. Specifically, sentence embedding encoded by the Long-Short Term Memory (LSTM) networks is concatenated to the noise vector as an input of the generator network, which produces a sequence of video frames by utilizing 3D convolutions. As such, temporal connections across frames are explicitly strengthened throughout the progress of video generation. In the discriminator network, in addition to determining whether videos are real or fake, the network must be capable of learning to align videos with the conditioning information. In particular, three discriminators are devised, including video discriminator, frame discriminator and motion discriminator. The former two classify realistic videos and frames from the generated ones, respectively, and also attempt to recognize the semantically matched video/frame-caption pairs from mismatched ones. The latter one is to distinguish the displacement between consecutive real or generated frames to further enhance temporal coherence. As a result, the whole architecture of TGANs-C is trained end-to-end by optimizing three losses, i.e., video-level and frame-level matching-aware loss to correct label of real or synthetic video/frames and align video/frames with correct caption, respectively, and temporal coherence loss to emphasize temporal consistency.

The main contribution of this work is the proposal of a new architecture, namely TGANs-C, which is one of the first effort towards generating videos conditioning on captions. This also leads to the elegant views of how to guarantee temporal coherence across generated video frames and how to align video/frame content with the given caption, which are the problems not yet fully understood in the literature. Through an extensive set of quantitative and qualitative experiments, we validate the effectiveness of our TGANs-C model on three different benchmarks.

\section{Related Work}\label{sec:RW}
We briefly group the related work into two categories: natural image synthesis and video generation. The former draws upon research in synthesizing realistic images by utilizing deep generative models, while the latter investigates generating image sequence/video from scratch.

\textbf{Image Synthesis.}
Synthesizing realistic images has been studied and analyzed widely in AI systems for characterizing the pixel level structure of natural images. There are two main directions on automatically image synthesis: Variational Auto-Encoders (VAEs) \cite{Kingma:ICLR13} and Generative Adversarial Networks (GANs) \cite{Goodfellow:NIPS14}. VAEs is a directed graphical model which firstly constrains the latent distribution of the data to come from prior normal distribution and then generates new samples through sampling from this distribution. This direction is straightforward to train but introduce potentially restrictive assumptions about approximate posterior distribution, always resulting in overly smoothed samples. Deep Recurrent Attentive Writer (DRAW) \cite{Karol:ICML15} is one of the early works which utilizes VAEs to generate images with a spatial attention mechanism. Furthermore, Mansimov \emph{et al.} extend this model to generate images conditioning on captions by iteratively drawing patches on a canvas and meanwhile attending to relevant words in the description \cite{Mansimov:ICLR16}.

GANs can be regarded as the generator network modules learnt with a two-player minimax game mechanism and has shown the distinct ability of producing plausible images \cite{Denton:NIPS15,Radford:ICLR16}. Goodfellow \emph{et al.} propose the theoretical framework of GANs and utilize GANs to generate images without any supervised information in \cite{Goodfellow:NIPS14}. Although the earlier GANs offer a distinct and promising direction for image synthesis, the results are somewhat noisy and blurry. Hence, Laplacian pyramid is further incorporated into GANs in \cite{Denton:NIPS15} to produce high quality images. Later in \cite{Odena2016conditional}, GANs is expended with a specialized cost function for classification, named auxiliary classifier GANs (AC-GANs), for generating synthetic images with global coherence and high diversity conditioning on class labels. Recently, Reed \emph{et al.} utilize GANs for image synthesis based on given text descriptions in \cite{Reed:ICML16}, enabling translation from character level to pixel level.

\textbf{Video Generation.}
When extending the existing generative models (e.g., VAEs and GANs) to video domain, very few works exploit such video generation from scratch task as both the spatial and temporal complex variations need to be characterized, making the problem very challenging. In the direction of VAEs, Mittal \emph{et al.} employ Recurrent VAEs and an attention mechanism in a hierarchical manner to create a temporally dependent image sequence conditioning on captions \cite{mittal2016sync}. For video generation with GANs, a spatio-temporal 3D deconvolutions based GANs is firstly proposed in \cite{vondrick2016generating} by untangling the scene's foreground from the background. Most recently, the 3D deconvolutions based GANs is further decomposed into temporal generator consisting of 1D deconvolutional layers and image generator with 2D deconvolutional layers for video generation in \cite{saito2016temporal}.

In short, our work in this paper belongs to video generation models capitalizing on adversarial learning. Unlike the aforementioned GANs-based approaches which mainly focus on video synthesis in an unconditioned manner, our research is fundamentally different in the way that we aim at generating videos conditioning on captions. In addition, we further improve video generation from the aspects of involving frame-level discriminator and strengthening temporal connections across frames.

\section{Video Generation from Captions}
The main goal of our Temporal GANs conditioning on Captions (TGANs-C) is to design a generative model with the ability of synthesizing a temporal coherent frame sequence semantically aligned with the given caption. The training of TGANs-C is performed by optimizing the generator network and discriminator network (video and frame discriminators which simultaneously judge synthetic or real and semantically mismatched or matched with the caption for video and frame) in a two-player minimax game mechanism. Moreover, the temporal coherence prior is additionally incorporated into TGANs-C to produce temporally coherent frame sequence in two different schemes. Therefore, the overall objective function of TGANs-C is composed of three components, i.e., video-level matching-aware loss to correct the label of real or synthetic video and align video with matched caption, frame-level matching-aware loss to further enhance the image reality and semantic alignment with the conditioning caption for each frame, and temporal coherence loss (i.e., temporal coherence constraint loss/temporal coherence adversarial loss) to exploit the temporal coherence between consecutive frames in unconditional/conditional scheme. The whole architecture of TGANs-C is illustrated in Figure \ref{fig:fig2}.

\subsection{Generative Adversarial Networks}
The basic generative adversarial networks (GANs) consists of two networks: a generator network $G$ that captures the data distribution for synthesizing image and a discriminator network $D$ that distinguishes real images from synthetic ones. In particular, the generator network $G$ takes a latent variable ${\bf{z}}$ randomly sampled from a normal distribution as input and produces a synthetic image ${x_{syn}} = G\left( {\bf{z}} \right)$. The discriminator network $D$ takes an image $x$ as input stochastically chosen (with equal probability) from real images or synthetic ones through $G$ and produces a probability distribution $P\left( {S|x} \right) = D\left( x \right)$ over the two image sources (i.e., synthetic or real). As proposed in \cite{Goodfellow:NIPS14}, the whole GANs can be trained in a two-player minimax game. Concretely, given an image example $x$, the discriminator network $D$ is trained to minimize the adversarial loss, i.e., maximizing the log-likelihood of assigning correct source to this example:
\begin{equation}\label{Eq:Eq1}\small
\begin{array}{l}
{l_a}(x) =  - {I_{(S = real)}}\log \big( {P(S = real|x)} \big)\\
\quad\quad\quad\quad~~ - (1 - {I_{(S = real)}})\log\big( {1 - P(S = real|x)} \big),
\end{array}
\end{equation}
where the indicator function $I_\emph{condition}=1$ if $\emph{condition}$ is true; otherwise $I_\emph{condition}=0$. Meanwhile, the generator network $G$ is trained to maximize the adversarial loss in Eq.(\ref{Eq:Eq1}), targeting for maximally fooling the discriminator network $D$ with its generated synthetic images $\{ {x}_{syn}\}$.

\subsection{Temporal GANs Conditioning on Captions (TGANs-C)}
In this section, we elaborate the architecture of our TGANs-C, the GANs based generative model consisting of two networks: a generator network $G$ for synthesizing videos conditioning on captions, and a discriminator network $D$ that simultaneously distinguishes real videos/frames from synthetic ones and aligns the input videos/frames with semantically matching captions. Moreover, two different schemes for modeling temporal coherence across frames are incorporated into TGANs-C for video generation.

\subsubsection{Generator Network}
Suppose we have an input sentence $\mathcal {S}$, where $\mathcal{S} = \{w_1, w_2, ..., w_{N_s-1}, w_{N_s}\}$ including $N_s$ words. Let ${\bf{w}}_t\in {{\mathbb{R}}^{d_w}}$ denote the $d_w$-dimensional ``one-hot" vector (binary index vector in a vocabulary) of the $t$-th word in sentence $\mathcal{S}$, thus the dimension of the textual feature ${\bf{w}}_t$, i.e., $d_w$, is the vocabulary size. Taking the inspiration from recent success of Recurrent Neural Networks (RNN) in image/video captioning \cite{pan2016jointly,pan2017video,xu2016msr,yao2017novel,yao2017boosting}, we first leverage the bidirectional LSTM (bi-LSTM) \cite{schuster1997bidirectional} to contextually embed each word and then encode the embedded word sequence into the sentence representation $\bf{S}$ via LSTM. In particular, the bi-LSTM consisting of forward and backward LSTMs \cite{Hochreiter:NC97} is adopted here. The forward LSTM reads the input word sequence in its natural order (from $w_1$ to $w_{N_s}$) and then calculates the forward hidden states sequence $\{\overrightarrow{h}_1, \overrightarrow{h}_2, ..., \overrightarrow{h}_{N_s}\}$, whereas the backward LSTM produces the backward hidden states sequence $\{\overleftarrow{h}_1, \overleftarrow{h}_2, ..., \overleftarrow{h}_{N_s}\}$ with the input sequence in the reverse order (from $w_{N_s}$ to $w_1$). The outputs of forward LSTM and backward LSTM are concatenated as the contextually embedded word sequence $\{h_1, h_2, ..., h_{N_s}\}$, where ${h_t} = \big[ {{\overrightarrow{h}_t}^\top,{\overleftarrow{h}_t}^\top} \big]^\top$. Then, we feed the embedded word sequence into the next LSTM-based encoder and treat the final LSTM output as the sentence representation ${\bf{S}} \in {{\mathbb{R}}^{d_s}}$. Note that both bi-LSTM and LSTM-based encoder are pre-learnt with sequence auto-encoder \cite{dai2015semi} in an unsupervised learning manner. Concretely, a LSTM-based decoder is additionally attached on the top of LSTM-based encoder for reconstructing the original word sequence. Such LSTM-based decoder will be removed and only the bi-LSTM and LSTM-based encoder are reserved for representing sentences with improved generalization ability after pre-training over large quantities of sentences.

Next, given the input sentence ${\bf{S}}$ and random noise variable ${\bf{z}}\in {\mathbb{R}}^{d_z} \sim \mathcal{N}(0,1)$, a generator network $G$ is devised to synthesize a frame sequence: $\{{{\mathbb{R}}^{{d_s}}} , {{\mathbb{R}}^{{d_z}}} \} \to {{\mathbb{R}}^{d_c \times d_l \times d_h \times d_d}}$ where $d_c$, $d_l$, $d_h$ and $d_d$ denote the channels number, sequence length, height and width of each frame, respectively. To model the spatio-temporal information within videos, the most natural way is to utilize the 3D convolutions filters \cite{tran2015learning} with deconvolutions \cite{zeiler2010deconvolutional} which can simultaneously synthesize the spatial information via 2D convolutions filters and provide temporal invariance across frames. Particularly, the generator network $G$ first encapsulates both the random noise variable ${\bf{z}}$ and input sentence ${\bf{S}}$ into a fixed-length input latent variable ${\bf{p}}$, which is applied with feature transformation and concatenation, and then synthesizes the corresponding video $v_{syn} = G\left( {{\bf{z}},{\bf{S}}} \right)$ based on the input ${\bf{p}}$ through 3D deconvolutional layers. The fixed-length input latent variable ${\bf{p}}$ is computed as
\begin{equation}\label{Eq:Eq3}\small
{\bf{p}} = \left[ {{{\bf{z}}}^\top,{\bf{S}}^\top{{\bf{W}}_s}} \right]^\top \in {{\mathbb{R}}^{d_z+d_p}},
\end{equation}
where ${{\bf{W}}_s} \in {{\mathbb{R}}^{{d_s} \times {d_{p}}}}$ is the transformation matrix for sentence representation. Accordingly, the generator network $G$ produces the synthetic video $v_{syn}=\{f^{1}_{syn}, f^{2}_{syn}, ..., f^{d_l}_{syn}\}$ conditioning on sentence $\mathcal {S}$ where $f^{i}_{syn} \in {{\mathbb{R}}^{d_c \times d_h \times d_d}}$ represents $i$-th synthetic frame.

\subsubsection{Discriminator Network}
The discriminator network $D$ is designed to enable three main abilities: (1) distinguishing real video from synthetic one and aligning video with the correct caption, (2) determining whether each frame is real/fake and semantically matched/mismatched with the conditioning caption, (3) exploiting the temporal coherence across consecutive real frames. To address the three crucial points, three basic discriminators are particularly devised:
\begin{itemize}

\item Video discriminator $D_0 \left( {v,\mathcal {S}} \right)$ ($\{{{\mathbb{R}}^{{d_v}}} , {{\mathbb{R}}^{d_s}} \} \to [0,1]$): $D_0$ first encodes input video $v \in {{\mathbb{R}}^{{d_v}}}$ into a video-level tensor ${\bf{m}}_v$ with a size of $d_{c_0} \times d_{l_0} \times d_{h_0} \times d_{d_0}$ via 3D convolutional layers. Then, the video-level tensor ${\bf{m}}_v$ is augmented with the conditioning caption ${\bf{S}}$ for discriminating whether the input video is real and simultaneously semantically matched with the given caption.

\item Frame discriminator $D_1 \left( {f^i,\mathcal {S}} \right)$ ($\{{{\mathbb{R}}^{{d_f}}} , {{\mathbb{R}}^{d_s}} \} \to [0,1]$): $D_1$ transforms each frame $f^i \in {{\mathbb{R}}^{{d_f}}}$ in $v$ into a frame-level tensor ${\bf{m}}_{f^i} \in {{\mathbb{R}}^{d_{c_0} \times d_{h_0} \times d_{d_0}}}$ through 2D convolutional layers and then augments frame-level tensor ${\bf{m}}_{f^i}$ with the conditioning caption ${\bf{S}}$ to recognize the real frames with matched caption.

\item Motion discriminator $D_2 \left( {f^i, f^{i-1}} \right)$ ($\{{{\mathbb{R}}^{{d_f}}} , {{\mathbb{R}}^{{d_f}}} \} \to {{\mathbb{R}}^{d_{c_0} \times d_{h_0} \times d_{d_0}}}$): $D_2$ distills the 2D motion tensor $\overrightarrow{{\bf{m}}}_{f^i}$ to represent the temporal dynamics across consecutive frames ${f^i}$ and $f^{i-1}$. Please note that we adopt the most direct way to measure such motion variance between two consecutive frames by subtracting previous frame-level tensor from current one (i.e., $\overrightarrow{{\bf{m}}}_{f^i}={\bf{m}}_{f^i}-{\bf{m}}_{f^{i-1}}$).
\end{itemize}

Specifically, in the training epoch, we can easily obtain a set of real-synthetic video triplets $\mathcal{T}$ according to the prior given captions, where each tuple $\{v_{syn^+}, v_{real^+}, v_{real^-}\}$ consists of one synthetic video $v_{syn^+}$ conditioning on given caption $\mathcal {S}$, one real video $v_{real^+}$ described by the same caption $\mathcal {S}$, and one real video $v_{real^-}$ described by different caption from $\mathcal {S}$. Therefore, three video-caption pairs are generated based on the caption $\mathcal {S}$ and its corresponding video tuple: the synthetic and semantically matched pair $\{v_{syn^+},\mathcal {S}\}$, real and semantically matched pair $\{v_{real^+},\mathcal {S}\}$, and another real but semantically mismatched pair $\{v_{real^-},\mathcal {S}\}$. Each video-caption pair $\{v,\mathcal {S}\}$ is then set as the input to the discriminator network $D$, followed by three kinds of losses to be optimized and each for one discriminator accordingly.

\textbf{Video-level matching-aware loss.} Noticing that the input video-caption pair $\{v,\mathcal {S}\}$ might not only be from distinctly sources (i.e., real or synthetic), but also contain matched or mismatched semantics. However, the conventional discriminator network can only differentiate the video sources without any explicit notion of the semantic relationship between video content and caption. Taking the inspiration from the matching-aware discriminator in \cite{Reed:ICML16}, we elaborate the video-level matching-aware loss for video discriminator $D_0$ to learn better alignment between video and the conditioning caption. In particular, for the video discriminator $D_0$, the conditioning caption $\mathcal {S}$ is first transformed with the embedding function ${\varphi _0}\left( {\bf{S}} \right) \in {{\mathbb{R}}^{d_{s_0}}}$ followed by rectification. Then the embedded sentence representation is spatially replicated to construct a $d_{s_0} \times d_{l_0} \times d_{h_0} \times d_{d_0}$ tensor, which is further concatenated with the video-level tensor ${\bf{m}}_v$ along the channel dimension. Finally the probability of recognizing real video with matched caption $D_0 \left( {v,\mathcal {S}} \right)$ is measured via a $1 \times 1 \times 1$ convolution followed by rectification and a $d_{l_0} \times d_{h_0} \times d_{d_0}$ convolution. Hence, given the real-synthetic video triplet $\{v_{syn^+}, v_{real^+}, v_{real^-}\}$ and the conditioning caption $\mathcal {S}$, the video-level matching-aware loss is measured as
\begin{equation}\label{Eq:Eq4}\small
\begin{array}{l}
{{\mathcal {L}}_v} =  - \frac{1}{3}\big[\log \left( {{D_0}\left( {{v_{rea{l^ + }}},\mathcal {S}} \right)} \right) + \log \left( {1 - {D_0}\left( {{v_{rea{l^ - }}},\mathcal {S}} \right)} \right)\\\\
\quad\quad\quad\quad\quad + \log ( {1 - {D_0}( {{v_{sy{n^ + }}},\mathcal {S}} )} )\big]
\end{array}.
\end{equation}
By minimizing this loss over positive video-caption pair (i.e., $\{v_{real^+},\mathcal {S}\}$) and negative video-caption pairs (i.e., $\{v_{syn^+},\mathcal {S}\}$ and $\{v_{real^-},\mathcal {S}\}$), the video discriminator $D_0$ is trained to not only recognize each real video from synthetic ones but also classify semantically matched video-caption pair from mismatched ones.

\textbf{Frame-level matching-aware loss.} To further enhance the frame reality and semantic alignment with the conditioning caption for each frame, a frame-level matching-aware loss is involved here which enforces the frame discriminator $D_1$ to discriminate whether each frame of the input video is both real and semantically matched with the caption. For the frame discriminator $D_1$, similar to $D_0$, an embedding function ${\varphi _1}\left( {\bf{S}} \right) \in {{\mathbb{R}}^{d_{s_0}}}$ is utilized to transform the conditioning caption $\mathcal {S}$ into the low-dimensional representation. Then we replicate the sentence embedding spatially to concatenate it with the frame-level tensor of each frame along the channel dimension. Accordingly, the final probability of recognizing real frame with matched caption $D_0 \left( {f^i,\mathcal {S}} \right)$ is achieved through a $1 \times 1$ convolution followed by rectification and a $d_{h_0} \times d_{d_0}$ convolution. Therefore, given the real-synthetic video triplet $\{v_{syn^+}, v_{real^+}, v_{real^-}\}$ and the conditioning caption $\mathcal {S}$, we calculate the frame-level matching-aware loss as
\begin{equation}\label{Eq:Eq5}\small
\begin{array}{l}
{{\mathcal {L}}_f} =  - \frac{1}{{3{d_l}}}\big[\sum\limits_{i = 1}^{{d_l}} {\log ( {{D_1}( {f_{rea{l^ + }}^i,\mathcal {S}} )} )}  + \sum\limits_{i = 1}^{{d_l}} {\log ( 1 - {{D_1}( {f_{rea{l^ - }}^i,\mathcal {S}} )} )} \\
\quad\quad\quad\quad\quad\quad + \sum\limits_{i = 1}^{{d_l}} {\log ( 1- {{D_1}( {f_{sy{n^ + }}^i,\mathcal {S}} )} )} \big]
\end{array},
\end{equation}
where $f_{rea{l^ + }}^i$, $f_{rea{l^ - }}^i$ and $f_{sy{n^ + }}^i$ denotes the $i$-th frame in $v_{rea{l^ + }}$, $v_{rea{l^ - }}$ and $v_{sy{n^ + }}$, respectively.

\textbf{Temporal coherence loss.} Temporal coherence is one generic prior for video modeling, which reveals the intrinsic characteristic of video that the consecutive video frames are usually visually and semantically coherent. To incorporate this temporal coherence prior into TGANs-C for video generation, we consider two kinds of schemes on the basis of motion discriminator $D_2 \left( {f^i, f^{i-1}}\right) $.

(1) \emph{Temporal coherence constraint loss.} Motivated by \cite{mobahi2009deep}, the similarity of two consecutive frames can be directly defined according to the Euclidean distances between their frame-level tensors, i.e., the magnitude of motion tensor:
\begin{equation}\label{Eq:Eq6}\small
{\mathcal{D}}\left( {{f^i},{f^{i - 1}}} \right) = \left\| {{{\bf{m}}_{{f^i}}} - {{\bf{m}}_{{f^{i - 1}}}}} \right\|_2^2 = \left\| {{\overrightarrow{{\bf{m}}}_{{f^i}}}} \right\|_2^2.
\end{equation}
Then, given the real-synthetic video triplet, we characterize the temporal coherence of the synthetic video $v_{syn^+}$ as a constraint loss by accumulating the Euclidean distances over every two consecutive frames:
\begin{equation}\label{Eq:Eq7}\small
{\mathcal{L}}_t^{(1)} = \frac{1}{{{d_l} - 1}}\sum\limits_{i = 2}^{{d_l}} {{\mathcal{D}}( {{f_{sy{n^ + }}^i},{f_{sy{n^ + }}^{i-1}}} )}.
\end{equation}
Please note that the temporal coherence constraint loss is designed only for optimizing generator network $G$. By minimizing this loss of synthetic video, the generator network $G$ is enforced to produce temporally coherent frame sequence.

(2) \emph{Temporal coherence adversarial loss.} Different from the first scheme formulating temporal coherence as a monotonous constraint in an unconditional manner, we further devise an adversarial loss to flexibly emphasize temporal consistency conditioning on the given caption. Similar to frame discriminator $D_1$, the motion tensor $\overrightarrow{{\bf{m}}}_{f^i}$ in motion discriminator $D_2$ is first augmented with embedded sentence representation ${\varphi _2}\left( {\bf{S}} \right)$. Next, such concatenated tensor representation is leveraged to measure the final probability ${\Phi _2}( {{\overrightarrow{{\bf{m}}}_{f^i}},\mathcal {S}} )$ of classifying the temporal dynamics between consecutive frames as real ones conditioning on the given caption. Thus, given the real-synthetic video triplet $\{v_{syn^+}, v_{real^+}, v_{real^-}\}$ and the conditioning caption $\mathcal {S}$, the temporal coherence adversarial loss is measured as
\begin{equation}\label{Eq:Eq8}\small
\begin{array}{l}
{\mathcal{L}}_t^{(2)} =  - \frac{1}{{3\left( {{d_l} - 1} \right)}}\big[\sum\limits_{i = 2}^{{d_l}} {\log ( {{\Phi _2}( {{\overrightarrow{{\bf{m}}}_{f_{rea{l^ + }}^i}},\mathcal {S}} )} )} \\
\quad\quad\quad\quad\quad\quad\quad+ \sum\limits_{i = 2}^{{d_l}} {\log ( {1 - {\Phi _2}( {{\overrightarrow{{\bf{m}}}_{f_{rea{l^ - }}^i}},\mathcal {S}} )} )} \\
\quad\quad\quad\quad\quad\quad\quad + \sum\limits_{i = 2}^{{d_l}} {\log ( {1 - {\Phi _2}( {{\overrightarrow{{\bf{m}}}_{f_{sy{n^ + }}^i}},\mathcal {S}} )} )} \big]
\end{array},
\end{equation}
where ${\overrightarrow{{\bf{m}}}_{f_{rea{l^ + }}^i}}$, ${\overrightarrow{{\bf{m}}}_{f_{rea{l^ - }}^i}}$ and ${\overrightarrow{{\bf{m}}}_{f_{sy{n^ + }}^i}}$ denotes the motion tensor in $v_{rea{l^ + }}$, $v_{rea{l^ - }}$ and $v_{sy{n^ + }}$, respectively. By minimizing the temporal coherence adversarial loss, the temporal discriminator $D_2$ is trained to not only recognize the temporal dynamics across synthetic frames from real ones but also align the temporal dynamics with the matched caption.

\subsubsection{Optimization}
The overall training objective function of TGANs-C integrates the video-level matching-aware loss in Eq.(\ref{Eq:Eq4}), frame-level matching-aware loss in Eq.(\ref{Eq:Eq5}) and temporal coherence constraint loss/temporal coherence adversarial loss in Eq.(\ref{Eq:Eq7})/Eq.(\ref{Eq:Eq8}). As our TGANs-C is a variant of the GANs architecture, we train the whole architecture in a two-player minimax game mechanism. For the discriminator network $D$, we update its parameters according to the following overall loss
\begin{equation}\label{Eq:Eq9}\scriptsize
{\mathcal{\hat L}^{(1)}_{D}} = \sum\limits_{\mathcal{T}}  {\frac{1}{2}\left( {{{\mathcal{L}}_v} + {{\mathcal{L}}_f}} \right)},
\end{equation}
\begin{equation}\label{Eq:Eq10}\scriptsize
{\mathcal{\hat L}^{(2)}_{D}} = \sum\limits_{\mathcal{T}}  {\frac{1}{3}\left( {{{\mathcal{L}}_v} + {{\mathcal{L}}_f} + {\mathcal{L}}_t^{(2)}} \right)},
\end{equation}
where $\mathcal{T}$ is the set of real-synthetic video triplets, ${\mathcal{\hat L}^{(1)}_{D}}$ and ${\mathcal{\hat L}^{(2)}_{D}}$ denotes the discriminator network $D$'s overall adversarial loss in unconditional scheme (i.e., TGANs-C with temporal coherence Constraint loss (TGANs-C-C)) and conditional scheme (i.e., TGANs-C with temporal coherence Adversarial loss (TGANs-C-A)), respectively. By minimizing this term, the discriminator network $D$ is trained to classify both videos and frames with correct sources, and simultaneously align videos and frames with semantically matching captions. Moreover, for TGANs-C-A, the discriminator network $D$ is additionally enforced to distinguish the temporal dynamics across frames with correct sources and also align the temporal dynamics with the matched captions.

For the generator network $G$, its parameters are adjusted with the following overall loss
\begin{equation}\label{Eq:Eq11}\scriptsize
\begin{array}{l}
{\mathcal{\hat L}^{(1)}_{G}} =  - \sum\limits_{{v_{sy{n^ + }}} \in {\mathcal{T}} } {} \frac{1}{3}\big[\log ( {{D_0}( {{v_{sy{n^ + }}},\mathcal {S}} )} ) + \frac{1}{{{d_l}}}\sum\limits_{i = 1}^{{d_l}} {\log ( {{D_1}( {f_{sy{n^ + }}^i,\mathcal {S}} )} )} \\
\quad\quad\quad\quad\quad\quad\quad\quad\quad\quad - \frac{1}{{{d_l} - 1}}\sum\limits_{i = 2}^{{d_l}} {{\mathcal{D}}( {{f_{sy{n^ + }}^i},{f_{sy{n^ + }}^{i-1}}} )}\big]
\end{array},
\end{equation}
\begin{equation}\label{Eq:Eq12}\scriptsize
\begin{array}{l}
{\mathcal{\hat L}^{(2)}_{G}} =  - \sum\limits_{{v_{sy{n^ + }}} \in {\mathcal{T}} } {} \frac{1}{3}\big[\log ( {{D_0}( {{v_{sy{n^ + }}},\mathcal {S}} )} ) + \frac{1}{{{d_l}}}\sum\limits_{i = 1}^{{d_l}} {\log ( {{D_1}( {f_{sy{n^ + }}^i,\mathcal {S}} )} )} \\
\quad\quad\quad\quad\quad\quad\quad\quad\quad\quad + \frac{1}{{{d_l} - 1}}\sum\limits_{i = 2}^{{d_l}} {\log ( {{\Phi _2}( {{\overrightarrow{{\bf{m}}}_{f_{sy{n^ + }}^i}},\mathcal {S}} )} )}\big]
\end{array},
\end{equation}
where ${\mathcal{\hat L}^{(1)}_{G}}$ and ${\mathcal{\hat L}^{(2)}_{G}}$ denotes the generator network $G$'s overall adversarial loss in TGANs-C-C and TGANs-C-A, respectively. The generator network $G$ is trained to fool the discriminator network $D$ on videos/frames source prediction with its synthetic videos/frames and meanwhile align synthetic videos/frames with the conditioning captions. Moreover, for TGANs-C-C, the consecutive synthetic frames are enforced to be similar in an unconditional scheme, while for TGANs-C-A, it additionally aims to fool $D$ on temporal dynamics source prediction with the synthetic videos in a conditional scheme. The training process of TGANs-C is given in Algorithm \ref{ag:ag01}.

\begin{algorithm}[!tb]\small
\caption{\small The training of Temporal GANs conditioning on Captions (TGANs-C)}\label{ag:ag01}
\begin{algorithmic}[1]
    \STATE
        Given the number of maximum training iteration $T$.

        \FOR{$t=1$ to $T$}
        \STATE
            Fetch input batch with sampled video-sentence pairs $\{(\mathcal {S},v_{real^+})\}$.
            \FOR{Each video-sentence pair $(\mathcal {S},v_{real^+})$}
            \STATE
                Get the random noise variable ${\bf{z}} \sim \mathcal{N}(0,1)$.
            \STATE
                Produce the synthetic video $v_{syn^+} = G\left( {{\bf{z}},{\bf{S}}} \right)$ conditioning on the caption $\mathcal {S}$ via the generator network $G$.
            \STATE
                Randomly select one real video $v_{real^-}$ described by a different caption from $\mathcal {S}$.
            \ENDFOR
            \STATE
                Obtain all the real-synthetic tuple $\{v_{syn^+}, v_{real^+}, v_{real^-}\}$ with the corresponding caption $\mathcal {S}$, denoted as $\mathcal{T}$ in total.
            \STATE
                Compute video-level matching-aware loss via Eq. (\ref{Eq:Eq4}).
            \STATE
                Compute frame-level matching-aware loss via Eq. (\ref{Eq:Eq5}).
            \STATE
                -Scheme 1: TGANs-C-C
            \STATE
                \hspace{7pt}Compute temporal coherence constraint loss via Eq. (\ref{Eq:Eq7}).
            \STATE
                \hspace{7pt}Update the discriminator network $D$ w.r.t loss in Eq. (\ref{Eq:Eq9}).
            \STATE
                \hspace{7pt}Update the generator network $G$ w.r.t loss in Eq. (\ref{Eq:Eq11}).
            \STATE
                -Scheme 2: TGANs-C-A
            \STATE
                \hspace{7pt}Compute temporal coherence adversarial loss via Eq. (\ref{Eq:Eq8}).
            \STATE
                \hspace{7pt}Update the discriminator network $D$ w.r.t loss in Eq. (\ref{Eq:Eq10}).
            \STATE
                \hspace{7pt}Update the generator network $G$ w.r.t loss in Eq. (\ref{Eq:Eq12}).
        \ENDFOR
\end{algorithmic}
\end{algorithm}

\subsection{Testing Epoch}
After the optimization of TGANs-C, we can obtain the learnt generator network $G$. Thus, given a test caption $\hat{\mathcal {S}}$, the bi-LSTM is first utilized to contextually embed the input word sequence, followed by a LSTM-based encoder to achieve the sentence representation $\hat{\bf{S}}$. The sentence representation $\hat{\bf{S}}$ is then concatenated with the random noise variable ${\bf{z}}$ as in Eq.(\ref{Eq:Eq3}) and finally fed into the generator network $G$ to produce the synthetic video $\hat v_{syn}=\{\hat f^{1}_{syn}, \hat f^{2}_{syn}, ..., \hat f^{d_l}_{syn}\}$.

\section{Experiments}\label{sec:EX}
We evaluate and compare our proposed TGANs-C with state-of-the-art approaches by conducting video generation task on three datasets of progressively increasing complexity: Single-Digit Bouncing MNIST GIFs (SBMG) \cite{mittal2016sync}, Two-digit Bouncing MNIST GIFs (TBMG) \cite{mittal2016sync}, and Microsoft Research Video Description Corpus (MSVD) \cite{Chen:ACL11}. The first two are recently released GIF-based datasets consisting of MNIST \cite{lecun1998gradient} digits moving frames and the last is a popular video captioning benchmark of YouTube videos.

\subsection{Datasets}
\begin{figure}[!tb]
    \centering {\includegraphics[width=0.5\textwidth]{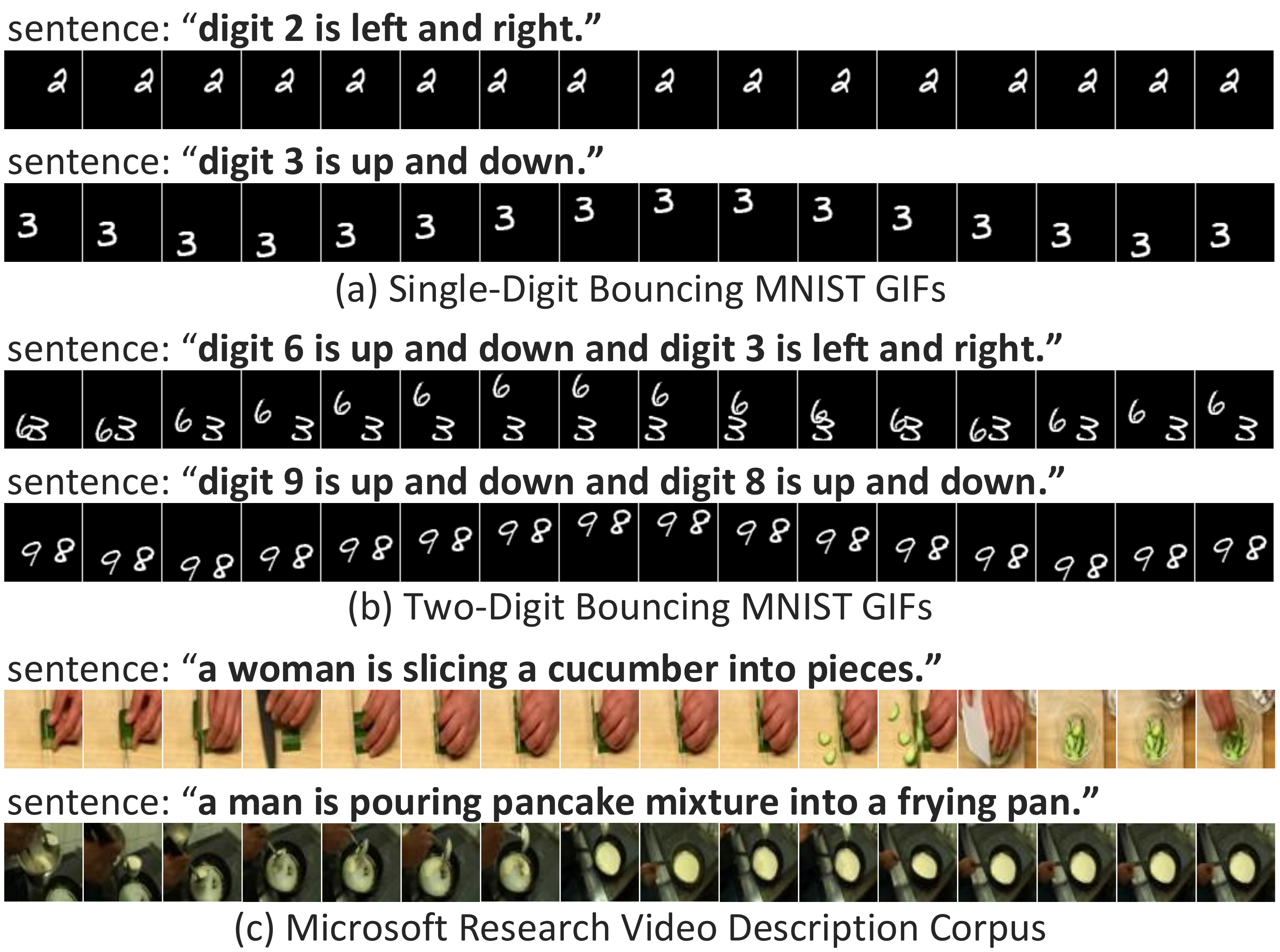}}
    \caption{\small (a)---(c): Exemplary video-caption pairs from three benchmarks: (a) Single-Digit Bouncing MNIST GIFs; (b) Two-Digit Bouncing MNIST GIFs; (c) Microsoft Research Video Description Corpus.}
    \label{fig:fig3}
\end{figure}

\textbf{SBMG.} Similar to priors works \cite{Srivastava:ICML15,shi2015convolutional} in generating synthetic dataset, SBMG is produced by having single handwritten digit bouncing inside a $64 \times 64$ frame. It is composed of 12,000 GIFs and every GIF is 16 frames long, which contains a single $28 \times 28$ digit moving left-right or up-down. The starting position of the digit is chosen uniformly at random. Each GIF is accompanied with single sentence describing the digit and its moving direction, as shown in Figure \ref{fig:fig3}(a).

\textbf{TBMG.} TBMG is an extended synthetic dataset of SBMG which contains two handwritten digits bouncing. The generation process is the same as SBMG and the two digits within each GIF move left-right or up-down separately. Figure \ref{fig:fig3}(b) shows two exemplary GIF-caption pairs in TBMG.

\textbf{MSVD.} MSVD contains 1,970 video snippets collected from YouTube. There are roughly 40 available English descriptions per video. In experiments, we manually filter out the videos about cooking and generate a subset of 518 cooking videos. Following the settings in \cite{Guadarrama:ICCV13}, our cooking subset is split with 363 videos for training and 155 for testing. Since video generation is a challenging problem, we assembled this subset with cooking scenario to better diagnose pros and cons of models. We randomly select two examples from this subset and show them in Figure \ref{fig:fig3}(c).

\subsection{Experimental Settings}
\textbf{Parameter Settings.}
We uniformly sample $d_l = 16$ frames for each GIF/video and each word in the sentence is represented as ``one-hot" vector. The architecture of our TGANs-C is mainly developed based on \cite{Radford:ICLR16,Reed:ICML16}. We resize all the GIFs/videos in three datasets with $48 \times 48$ pixels. In particular, for sentence encoding, the dimension of the input and hidden layers in bi-LSTM and LSTM-based encoder are all set to 256. For the generator network $G$, the dimension of random noise variable ${\bf{z}}$ is 100 and the dimension of sentence embedding in generator network $d_p$ is 256. For the discriminator network $D$, we set the size of video-level tensor ${\bf{m}}_v$ in video discriminator $D_{0}$ as $512 \times 1 \times 3 \times 3$ and the size of frame-level tensor ${\bf{m}}_{f^i}$ in frame discriminator $D_{1}$ is as $512 \times 3 \times 3$.

\textbf{Implementation Details.}
We mainly implement our proposed method based on Theano \cite{al2016theano}, which is one of widely adopted deep learning frameworks. Following the standard settings in \cite{Radford:ICLR16}, we train our TGANs-C models on all datasets by utilizing Adam optimizer with a mini-batch size of 64. All weights were initialized from a zero-centered Normal distribution with standard deviation 0.02 and the slope of the leak was set to 0.2 in the LeakyReLU. We set the learning rate and momentum as 0.0002 and 0.9, respectively.

\textbf{Evaluation Metric.}
For the quantitative evaluation of video generation, we adopt Generative Adversarial Metric (GAM) \cite{im2016generating} which can directly compare two generative adversarial models by having them engage in a ``battle" against each other. Given two generative adversarial models $M_1=\{(\widetilde{G}_1,\widetilde{D}_1)\}$ and $M_2=\{(\widetilde{G}_2,\widetilde{D}_2)\}$, two kinds of ratios between the discriminative scores of the two models are measured as:
\begin{equation}\label{Eq:Eq13}\small
{r_{test}} = \frac{{\epsilon \left( {{{\widetilde D}_1}\left( {{{\bf{x}}_{test}}} \right)} \right)}}{{\epsilon \left( {{{\widetilde D}_2}\left( {{{\bf{x}}_{test}}} \right)} \right)}}~~~~~{\rm{and}}~~~~~{r_{sample}} = \frac{{\epsilon \left( {{{\widetilde D}_1}\left( {{\widetilde{G}_2}\left( {\bf{z}} \right)} \right)} \right)}}{{\epsilon \left( {{{\widetilde D}_2}\left( {{\widetilde{G}_1}\left( {\bf{z}} \right)} \right)} \right)}},
\end{equation}
where $\epsilon \left(  \bullet  \right)$ denotes the classification error rate and ${\bf{x}}_{test}$ is the testing set. The test ratio ${r_{test}}$ shows which model generalizes better on test data and the sample ratio ${r_{sample}}$ reveals which model can fool the other model more easily. Finally, the GAM evaluation metric judges the winner as:
\begin{equation}\label{Eq:Eq13}\small
{\rm{winner}} = \left\{ \begin{array}{l}
{M_1}\hspace{6pt}{\rm{if}}~~{r_{sample}} < 1~{\rm{and}}~{r_{test}}\simeq 1\\
{M_2}\hspace{6pt}{\rm{if}}~~{r_{sample}} > 1~{\rm{and}}~{r_{test}}\simeq 1\\
{\rm{Tie}}\hspace{7pt}{\rm{otherwise}}
\end{array} \right..
\end{equation}

\subsection{Compared Approaches}
To empirically verify the merit of our TGANs-C, we compared the following state-of-the-art methods.

(1) Synchronized Deep Recurrent Attentive Writer (Sync-DRAW) \cite{mittal2016sync}: Sync-DRAW is a VAEs-based model for video generation conditioning on captions which utilizes Recurrent VAEs to model spatio-temporal relationship and a separate attention mechanism to capture local saliency.

(2) Generative Adversarial Network for Video (VGAN) \cite{vondrick2016generating}: The original VGAN attempts to leverage the spatio-temporal convolutional architecture to design a GANs-based generative model for video generation in an unconditioned manner. Here we additionally incorporate the matching-aware loss into the discriminator network of basic VGAN and enable this baseline to generate videos conditioning on captions.

(3) Generative Adversarial Network with Character-Level Sentence encoder (GAN-CLS) \cite{Reed:ICML16}: GAN-CLS is originally designed for image synthesis from text descriptions by utilizing DC-GAN and a hybrid character-level convolutional-recurrent neural network for text encoding. We directly extend this architecture by replacing 2D convolutions with 3D spatio-temporal convolutions for text-conditional video synthesis.

(4) Temporal GANs conditioning on Captions (TGANs-C) is our proposal in this paper which includes two runs in different schemes: TGANs-C with temporal coherence constraint loss (TGANs-C-C) and TGANs-C with temporal coherence adversarial loss (TGANs-C-A). Two slightly different settings of TGANs-C are named as TGANs-C$_1$ and TGANs-C$_2$. The former is trained with only video-level matching-aware loss, while the latter is more similar to TGANs-C that only excludes the temporal coherence loss.

\begin{figure}[!tb]
    \centering {\includegraphics[width=0.46\textwidth]{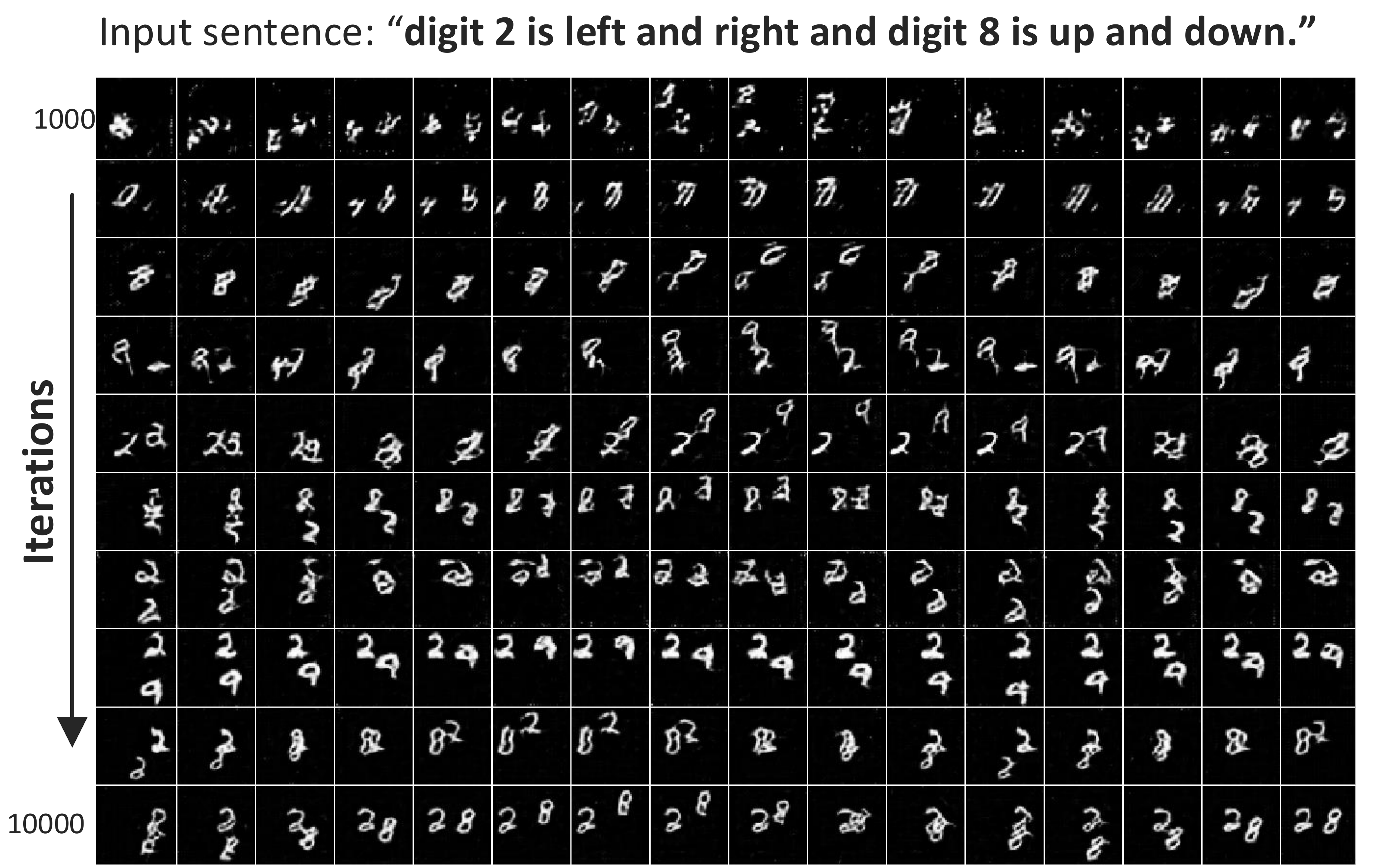}}
    \vspace{-0.1in}
    \caption{\small Evolution of synthetic results of the generator network $G$ with the increase of the iteration on TBMG dataset. Both of the input random noise variable ${\bf{z}}$ and caption $\mathcal {S}$ are fixed. Each row denotes one synthetic video and the results are shown every 1,000 iterations.}
    \label{fig:figop}
    \vspace{-0.2in}
\end{figure}

\begin{figure*}[!tb]
   \centering {\includegraphics[width=0.83\textwidth]{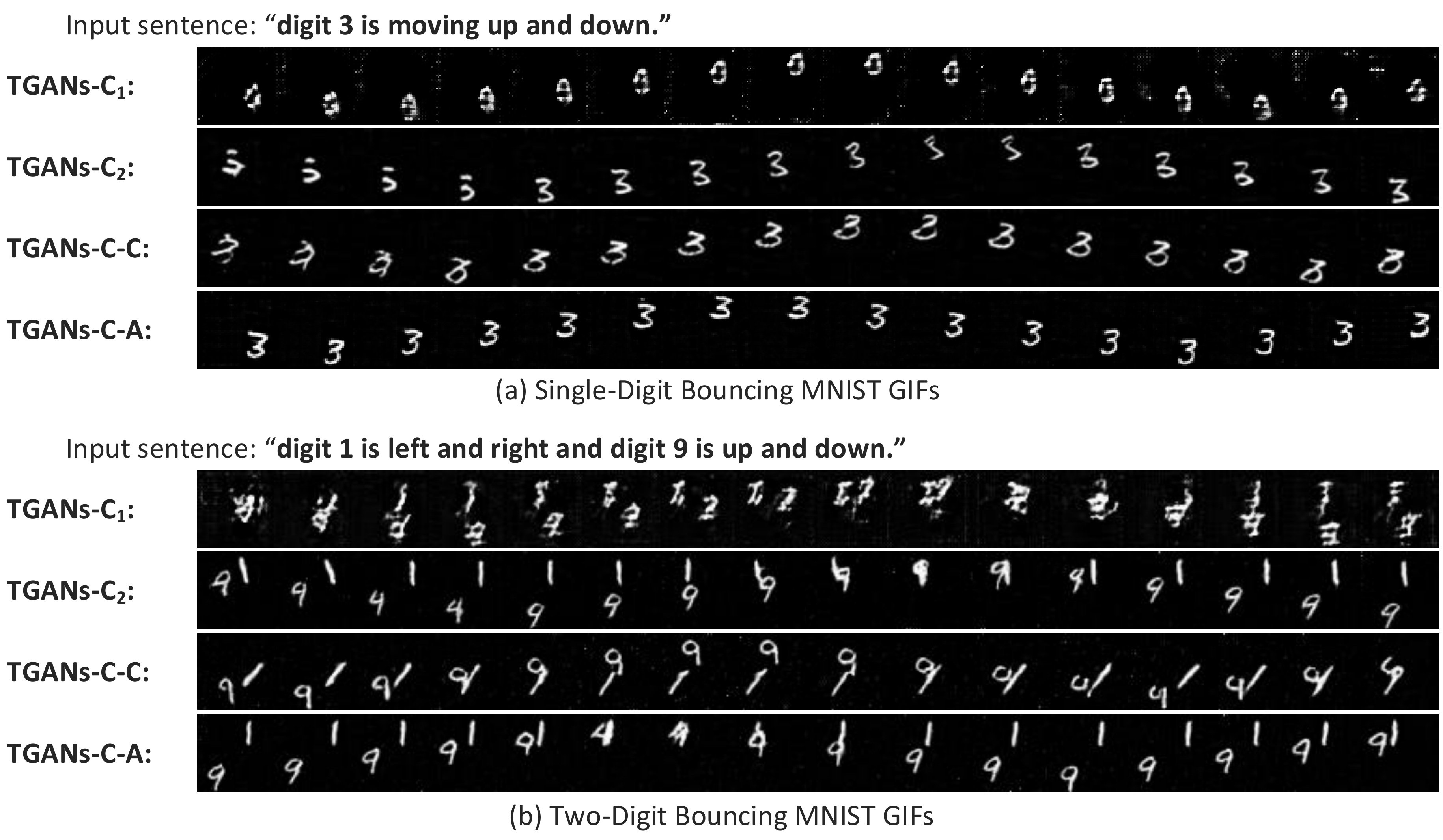}}
   \vspace{-0.15in}
   \caption{\small Examples of generated videos by our four TGANs-C runs on (a) Single-Digit Bouncing MNIST GIFs and (b) Two-digit Bouncing MNIST GIFs.}
   \label{fig:fig4}
   \vspace{-0.15in}
\end{figure*}

\subsection{Optimization Analysis}
Different from the traditional discriminative models which have a particularly well-behaved gradient, our TGANs-C is optimized with a complex two-player minimax game. Hence, we depict the evolution of the generator network $G$ at the training stage to illustrate the convergence of our TGANs-C. Concretely, we randomly sample one random noise variable ${\bf{z}}$ and caption $\mathcal {S}$ before training, and then leverage them to produce synthetic videos via the generator networks $G$ of TGANs-C-A at different iterations on TBMG. As shown in Figure \ref{fig:figop}, the quality of synthetic videos does improve as the iterations increase. Specifically, after 9,000 iterations, the generater network $G$ consistently synthesizes plausible videos by reproducing the visual appearances and temporal dynamics of handwritten digits conditioning on the caption.

\subsection{Qualitative Evaluation}
We then visually examine the quality of the results and compare among our four internal TGANs-C runs on SBMG and TBMG datasets. The examples of generated videos are shown in Figure \ref{fig:fig4}. Given the input sentence of ``digit 3 is moving up and down" in Figure \ref{fig:fig4}(a), all the four runs can interpret the temporal track of forming single-digit bouncing videos. TGANs-C$_1$ which only judges real or fake on video level and aligns video with the caption performs the worst among all the models and the predicted frames tend to be blurry. By additionally distinguishing frame-level realness and optimizing frame-caption matching, TGANs-C$_2$ is capable of producing videos in which each frame is clear but the shape of the digit sometimes changes over time. Compared to TGANs-C$_2$, TGANs-C-C emphasizes the coherence across adjacent frames by further regularizing the similarity in between. As a result, the frames generated by TGANs-C-C are more consistent than TGANs-C$_2$ particularly of the digit in the frames, but on the other hand, the temporal coherence constraint exploited in TGANs-C-C is in a brute-force manner, making the generated videos monotonous and not that real. TGANs-C-A, in comparison, is benefited from the mechanism of adversarially modeling temporal connections. The chance that a video is gradually formed as real is better.

\begin{figure*}[!tb]
   \centering {\includegraphics[width=0.85\textwidth]{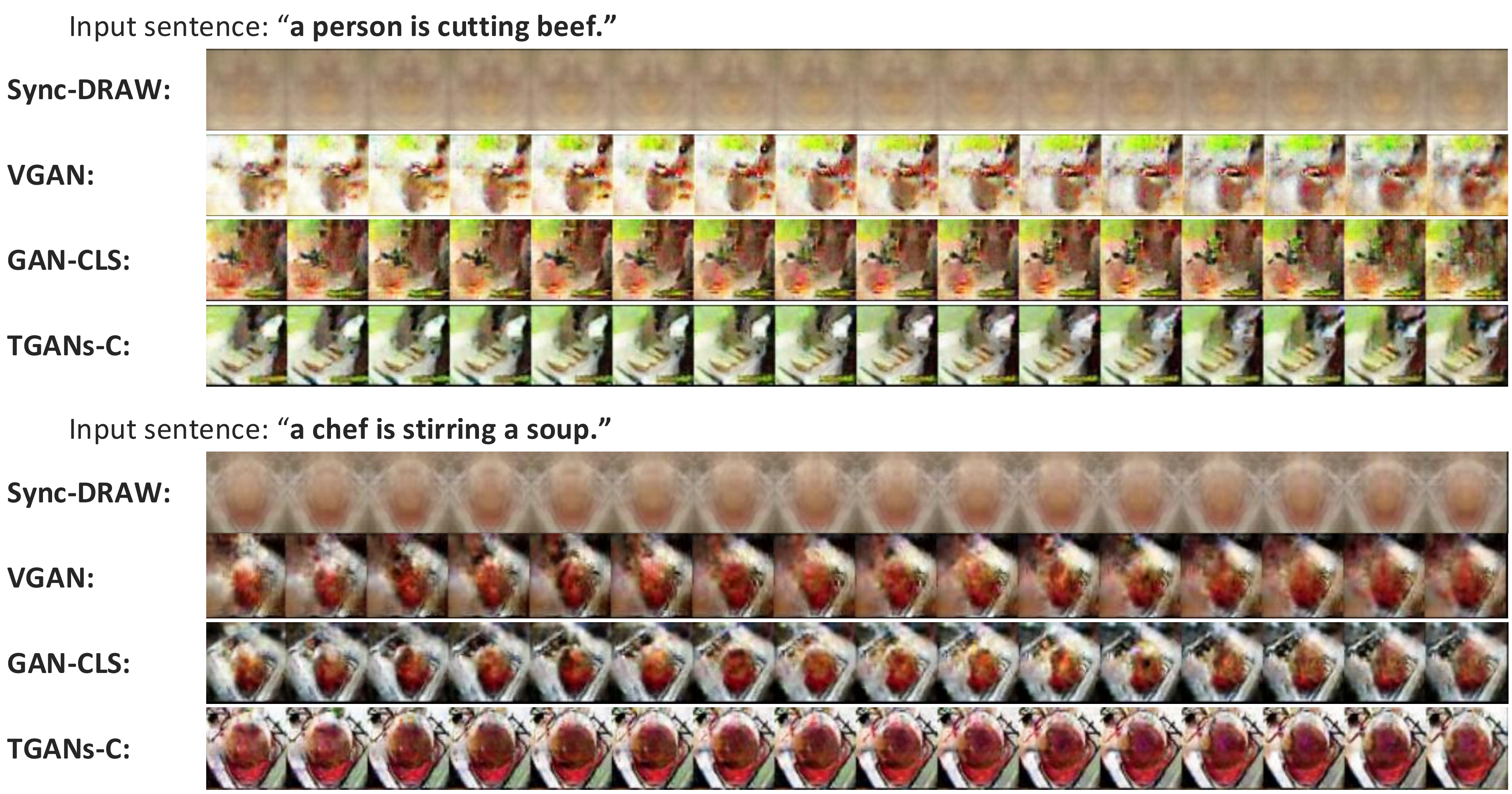}}
   \vspace{-0.15in}
   \caption{\small Examples of generated videos by different approaches on MSVD dataset.}
   \label{fig:fig5}
   \vspace{-0.2in}
\end{figure*}

Figure \ref{fig:fig4}(b) shows the generated videos by our four TGANs-C runs conditioning on the caption of ``digit 1 is left and right and digit 9 is up and down." Similar to the observations on single-digit bouncing videos, the four runs could also model the temporal dynamics of two-digit bouncing scenarios. When taking temporal smoothness into account, the quality of the videos generated by TGANs-C-C and TGANs-C-A is enhanced, as compared to the videos produced by TGANs-C$_1$ and TGANs-C$_2$. In addition, TGANs-C-A generates more realistic videos than TGANs-C-C, verifying the effectiveness of learning temporal coherence in an adversarial fashion.

Next, we compare with the three baselines on MSVD dataset. In view that TGANs-C-A consistently performs the best in our internal comparisons, we refer to this run as TGANs-C in the following evaluations. The comparisons of generated videos by different approaches are shown in Figure \ref{fig:fig5}. We can easily observe that the videos generated by our TGANs-C have higher quality compared to the other models. The created frames by Sync-DRAW are very blurry since VAEs are biased towards generating smooth frames and the method does not present all the objects in the frames. The approach of VGAN generates the frames which tend to be fairly sharp. However, the background of the frames is stationary as VGAN enforces a static background and moving foreground, making it vulnerable to produce videos with background movement. Compared to GAN-CLS which only involves video-level matching-aware discriminator, our TGANs-C takes the advantages of additionally exploring frame-level matching-aware discriminator and temporal coherence across frames, and thus generates more realistic~videos.

\subsection{Human Evaluation}
To better understand how satisfactory are the videos generated from different methods, we also conducted a human study to compare our TGANs-C against three approaches, i.e., Sync-DRAW, VGAN and GAN-CLS. A total number of 30 evaluators (15 females and 15 males) from different education backgrounds, including computer science (8), management (4), business (4), linguistics (4), physical education (1), international trade (1) and engineering (8), are invited and a subset of 500 sentences is randomly selected from testing set of MSVD dataset for the subjective evaluation.

We show all the evaluators the four videos generated by each approach plus the given caption and ask them to rank all the videos from 1 to 4 (good to bad) with respect to the three criteria: 1) Reality: how realistic are these generated videos? 2) Relevance: whether the videos are relevant to the given caption? 3) Coherence: judge the temporal connection and readability of the videos. To make the annotation as objective as possible, the four generated videos conditioning on each sentence are assigned to three evaluators and the final ranking is averaged on the three annotations. Furthermore, we average the ranking on each criterion of all the generated videos by each method and obtain three metrics. Table \ref{table:human-msvd} lists the results of the user study on MSVD dataset. Overall, our TGANs-C is clearly the winner across all the three criteria.

\begin{table}[]\small
\centering
\caption{\small The user study on three criteria: 1) Reality - how realistic are these generated videos? 2) Relevance - whether the videos are relevant to the given caption? 3) Coherence - judge the temporal connection and readability of the videos. The average ranking (lower is better) on each criterion of all the generated videos by each approach is reported.}
\vspace{-0.15in}
\label{table:human-msvd}
\begin{tabular}{l|ccc}
\hline
\textbf{Methods} & \textbf{Reality} & \textbf{Relevance} & \textbf{Coherence} \\ \hline
Sync-DRAW        & 3.95             & 3.93               & 3.90               \\
VGAN             & 2.21             & 2.29               & 2.23               \\
GAN-CLS          & 2.08             & 1.97               & 2.01               \\ \hline
TGANs-C          & \textbf{1.76}    & \textbf{1.81}      & \textbf{1.86}      \\ \hline
\end{tabular}
\vspace{-0.2in}
\end{table}

\subsection{Quantitative Evaluation}
To further quantitatively verify the effectiveness of our proposed model, we compare our TGANs-C with two generative adversarial baselines (i.e., VGAN and GAN-CLS) in terms of GAM evaluation metric on MSVD dataset. As the method of Sync-DRAW produces videos by VAEs-based architecture rather than generative adversarial scheme, it is excluded in this comparison. The quantitative results are summarized in Table \ref{table:GAM}. Overall, considering the ``battle" between our TGANs-C and the other two baselines, the sample ratios ${r_{sample}}$ are both less than one, indicating that TGANs-C can produce more authentic synthetic videos and fool the other two models more easily. The results basically verify the advantages of exploiting frame-level realness, frame-caption matching and the temporal coherence across adjacent frames for video generation. Moreover, when comparing between the two 3D-based baselines, GAN-CLS beats VGAN easily. This somewhat reveals the weakness of VGAN, where the architecture is devised with the brute-force assumption that the background is stationary and only foreground moves, making it hard to mimic the real-word videos with dynamic background. Another important observation is that for the ``battle" between each two runs, the test ratio ${r_{test}}$ is consistently approximately equal to one. This assures that none of the discriminator networks $D$ in these runs is overfitted more than the other, i.e., the corresponding sample ratios ${r_{sample}}$ are applicable and not biased for evaluating generative adversarial~models.

\begin{table}[!tb]\small
\centering
\caption{\small Model Evaluation with GAM metric on MSVD.}
\vspace{-0.15in}
\label{table:GAM}
\begin{tabular}{l|cc|c}
\hline
\textbf{Battler}        & ${\bf{r_{test}}}$ & ${\bf{r_{sample}}}$ & \textbf{Winner} \\ \hline
GAN-CLS vs VGAN      & 1.08           & 0.89             & GAN-CLS \\ \hline
TGANs-C vs VGAN       & 1.09           & 0.39             & TGANs-C       \\ \hline
TGANs-C vs GAN-CLS & 0.96           & 0.53             & TGANs-C       \\ \hline
\end{tabular}
\vspace{-0.2in}
\end{table}

\section{Conclusions}
Synthesizing images or videos will be crucial for the next generation of multimedia systems. In this paper, we have presented the Temporal GANs conditioning on Captions (TGANs-C) architecture, succeeded in generating videos that correspond to a given input caption. Our model expands on adversarial learning paradigm from three aspects. First, we extend 2D generator network to 3D for explicitly modeling spatio-temporal connections in videos. Second, in addition to naive discriminator network which only judges fake or real, ours further evaluate whether the generated videos or frames match the conditioning caption. Finally, to guarantee the adjacent frames coherently formed over time, the motion information between consecutive real or generated frames is taken into account in the discriminator network. Extensive quantitative and qualitative experiments conducted on three datasets validate our proposal and analysis. Moreover, our approach creates videos with better quality by a user study from 30 human subjects.

Future works will focus, first of all, on improving visual discriminability of our model, i.e., synthesize higher resolution videos. A promising route to explore will be that of decomposing the problem into several stages, where the shape or basic color based on the given caption is sketched in the primary stages and the advanced stages rectify the details of videos. Second, how to generate videos conditioning on open-vocabulary caption is expected. Last but not least, extending our framework to audio domain should be also interesting.

\textbf{Acknowledgments.} This work was supported in part by 973 Program under contract No. 2015CB351803 and NSFC under contract No. 61325009.

\bibliographystyle{ACM-Reference-Format}
\bibliography{sigproc}


\end{document}